\documentclass[letterpaper, 10 pt, conference]{ieeeconf}  

\IEEEoverridecommandlockouts                              

\usepackage{cite}
\usepackage{amsmath,amssymb,amsfonts}
\usepackage{algorithmic}
\usepackage{cuted}
\usepackage{multirow}
\usepackage{longtable}
\usepackage[export]{adjustbox}
\usepackage{graphicx}
\usepackage{lipsum}
\usepackage{textcomp}
\usepackage{xcolor}
\usepackage{siunitx}
\usepackage{float}
\usepackage[caption=false]{subfig}
\usepackage[bottom]{footmisc}
\usepackage[utf8]{inputenc}
\usepackage[T1]{fontenc}
\usepackage{booktabs}
\usepackage{tabularray}
\usepackage{gensymb}

\def\BibTeX{{\rm B\kern-.05em{\sc i\kern-.025em b}\kern-.08em
    T\kern-.1667em\lower.7ex\hbox{E}\kern-.125emX}}

\title{\LARGE \bf
Bistable SMA-driven engine for pulse-jet locomotion in soft aquatic robots
}

\author{Graziella Bedenik$^{1\ast}$, Antonio Morales$^{1\ast}$, Supun Pieris$^{1}$, Barbara da Silva$^{1}$, John W. Kurelek$^{1}$, \\ Melissa Greeff$^{2}$, and Matthew Robertson$^{1}$
\thanks{*These two authors contributed equally to this work.}
\thanks{$^{1}$Mechanical and Materials Engineering Department, Queen's University, Kingston, ON K7L 2V9, Canada.}%
\thanks{$^{2}$Electrical and Computer Engineering Department, Queen's University, Kingston, ON K7L 3N9, Canada.
        {\tt\small Correspondence: graziella.bedenik@queensu.ca}}%
}

\begin{document}

\maketitle
\thispagestyle{empty}
\pagestyle{empty}

\begin{abstract}
This paper presents the design and experimental validation of a bio-inspired soft aquatic robot, the DilBot, which uses a bistable shape memory alloy-driven engine for pulse-jet locomotion. Drawing inspiration from the efficient swimming mechanisms of box jellyfish, the DilBot incorporates antagonistic shape memory alloy springs encapsulated in silicone insulation to achieve high-power propulsion. The innovative bistable mechanism allows continuous swimming cycles by storing and releasing energy in a controlled manner. Through free-swimming experiments and force characterization tests, we evaluated the DilBot's performance, achieving a peak speed of 158 mm/s and generating a maximum thrust of 5.59 N. This work demonstrates a novel approach to enhancing the efficiency of shape memory alloy actuators in aquatic environments. It presents a promising pathway for future applications in underwater environmental monitoring using robotic swarms.


\end{abstract}

\section{Introduction}
\label{sec:intro}



The use of aquatic robots has risen in the last decades \cite{schweim2024unmanned,hasan2024oceanic}. This increase leverages advancements in both sensing and actuation devices, leading to more cost-effective and compact robotic solutions with greater ranges of operation and extended lifetime \cite{kazemi2019robotic}.


For tasks like environmental monitoring, samples are traditionally collected manually, with the help of extensively equipped boats, remotely operated vehicles (ROVs) \cite{berlian2016design}, static sensor networks \cite{adu2017water}, or divers that survey a point of interest \cite{national2016spills}. These approaches are costly both financially and in requiring specialized operatives. They are also often unsafe for researchers and involved staff and yield primarily offline and stationary measurements \cite{thenius2018subcultron,osterloh2012monsun,isokeit2017cooperative}.

\begin{figure}[t!]
    \centering
    \includegraphics[width=1\linewidth]{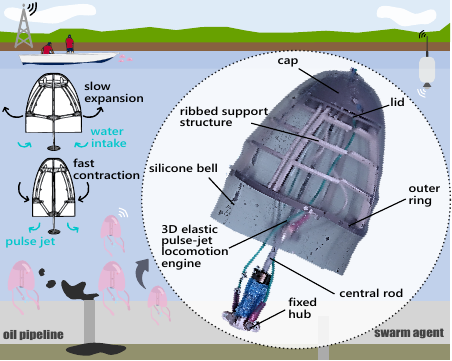}
    \caption{The morphology and function of a biological box jellyfish are reflected in the DilBot, highlighted in the circle on the right. It consists mainly of a silicone bell attached to an SMA-driven engine. The robot and animal both achieve swimming by cyclic opening and closing of their bells to generate water jet thrust for propulsion. Water is slowly drawn into the body's interior when the robot bell expands. The water is then forced outwards upon rapid bell contraction to produce a reaction force that facilitates the net forward motion.}
    \label{fig:swarm}
\end{figure}

As a new alternative, autonomous underwater vehicles (AUVs) are envisioned as a key future technology for underwater autonomous investigations \cite{meyer2017using,hoeher2021underwater,chen2022risk}, providing the needed data and saving costs by reducing the workforce and related safety risks \cite{osterloh2012monsun,isokeit2017cooperative}. Coordinated swarms of small and reasonably priced AUVs can potentially detect rapidly changing conditions in challenging environments,  such as the volume of oil in a spill site \cite{meyer2017using,isokeit2017cooperative,osterloh2012monsun,saleh2024situ}. 






Due to the challenging nature of the underwater environment, researchers often draw inspiration from nature to recreate bio-inspired behaviours for robots in search of optimized locomotion performance \cite{kazemi2019robotic}. The species under the sub-phylum Medusozoa, commonly called jellyfish, are often the subject of such research endeavours \cite{villanueva_jellyfish_2009,wang_design_2023,frame_thrust_2018,kazemi2019robotic,ye_lm-jelly_2022,almubarak_kryptojelly_2020,nir_jellyfish-like_2012,matharu_jelly-z_2023
}. They are not only one of the structurally simplest animals but also known to be one of the most efficient ocean swimmers, spending less energy per distance than many other underwater creatures due to recapturing part of the expended energy of each stroke to continue moving \cite{kazemi2019robotic,villanueva_jellyfish_2009,miles2019naut}.


Nevertheless, recreating the movement of an underwater creature while following traditional robot design strategies, such as off-the-shelf actuators and primarily rigid mechanical components, limits recreating the capabilities of biological counterparts. Incorporating soft and smart materials into robot design helps obtain the mimicry effect because they can perform flexible and complex movements. These materials also give an advantage when fabricating robots with a relatively small weight and volume compared to those manufactured using conventional methods and devices \cite{chu2012review}. 

Shape memory alloy (SMA) actuators are commonly used for this purpose \cite{luo2024design,li2022design,sapmaz2021nickel,kazemi2019robotic,almubarak_kryptojelly_2020,chu2012review,Villaneuva2011,cruz2020soft}, especially in spring form to allow more significant displacement, due to large driving strain, high working density, and quiet and low driving voltage \cite{wang2023soft}. These work by passing an electric current through the SMA wires to thermally induce a shape memory effect, which overcomes the spring force and causes the SMA wires to contract. However, they present a significant drawback, a slow actuation rate due to heat transfer considerations \cite{chu2012review,kazemi2019robotic}. For this reason, they are not usually applied to produce high-power motions under joule heating due to their high energy consumption, which could be exacerbated underwater due to cooler surrounding temperatures. Using bistable mechanisms, which can quickly store and release high-power energy, even in aquatic environments, could make these actuators more efficient for pulse-jet locomotion \cite{chen2018harnessing,bambrick2024does,ali2022novel}. However, the combined use of SMA-driven bistable mechanisms has not yet been explored in the literature for pulse jet locomotion underwater.



For instance, in \cite{sapmaz2021nickel}, the authors present a caterpillar-like prototype with two SMA spring actuators integrated into the soft robot using custom structures to emulate a crawling motion inside fluid pipelines. The coils received no insulating treatment and could only work above 24\degree C water. In \cite{luo2024design},  SMA springs drive the tentacles of a jellyfish-inspired robot. By controlling the power-on time of the actuators, it undergoes varying degrees of deformation, squeezing out the water inside the robot and forming a central jet effect. Despite the ability to produce both linear and angular motion through pulse-jet locomotion, there are no gimmicks to improve the material's cooling/heating ratio time, resulting in a consumption of up to 10 A per actuator. Meanwhile, in  \cite{kazemi2019robotic} and \cite{cruz2020soft}, the authors of both works present jellyfish-inspired underwater robots where the actuators go through some encapsulating process. The necessary current to actuate the robots was smaller, up to 6 A. 

All the aforementioned discussed works share the need to wait for the spring actuators to ``reset'' and return to their original state before the next cycle due to the SMA characteristic. In contrast, in \cite{li2022design}, the authors propose a turtle-inspired robot driven by an antagonistic SMA spring mechanism. It compensates for the resetting drawback, enabling higher speeds and actuating rates. Nevertheless, no experiment was performed underwater. 

To address these gaps in the current literature, this work describes a bioinspired robot, the DilBot. It leverages the swimming principle employed by some Meduzoa organisms for underwater locomotion. Emulating natural animal behaviour, the proof-of-concept (POC) prototype illustrated in Fig. \ref{fig:swarm} is driven by antagonistic SMA springs in a symmetric and bistable engine arrangement for rapid swimming propulsion. The mechanism generates cyclic opening and closing of a thin, stretchable membrane, the bell that enables asymmetric impulse generation from water jet thrust for net forward
motion.

This design circumvents the traditional limitation of low SMA spring actuator bandwidth, which typically restricts the speed and power of the robot's motion. It allows the generation of cyclic high-power motion even during the reset phase of the SMA actuator pairs. Another contribution of this work is the manufacturing process of novel encapsulated SMA actuators called SpIRAl (silicone-insulated robotic actuator) that provide thermal and electrical insulation to increase the efficiency of powering the springs underwater. Unlike the state-of-the-art, SpiRAls are not cast directly into silicone moulds \cite{wang2009micro,wang_design_2023,kazemi2019robotic,Villaneuva2011} or stuffed entirely into insulating tubes \cite{cruz2020soft,almubarak_kryptojelly_2020}. This feature allows for easy maintenance and reconfiguration of the actuators, the robot, and its design, all while maintaining the desired insulating properties for the application.

\section{Design}
\label{sec:system}

The two main classes of jellyfish are Scyphozoa, or true jellyfish, and Cubozoa, or box jellyfish. The first use their rounder bell to combine jet propulsion with rowing and paddling motions, while the latter has a bullet-like bell used to swim through jet propulsion only \cite{chu2012review,kazemi2019robotic}. Since box jellyfish exclusively produce high-power, strong jets to move, their motion is ambitious to replicate with SMA springs, a challenge we address in this work through developing our POC prototype. The pertinent design aspects of the robot are broken down into subsections \ref{subsec:components},  \ref{subsec:engine}, and \ref{subsec:locomotion}.

\subsection{Body Components}
\label{subsec:components}

\begin{figure*}[ht!] 
    \centering
  \subfloat[\label{design_CAD}]{%
       \includegraphics[height=0.33\linewidth]{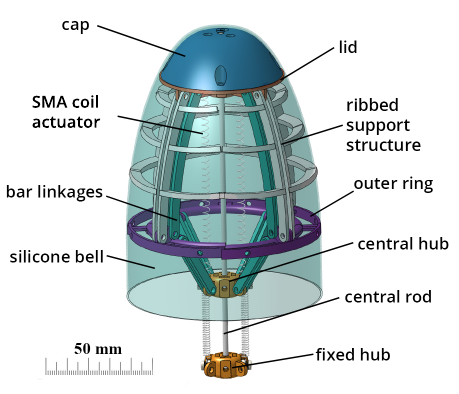}}
    \hfill
  \subfloat[\label{design_concept}]{%
        \includegraphics[height=0.33\linewidth]{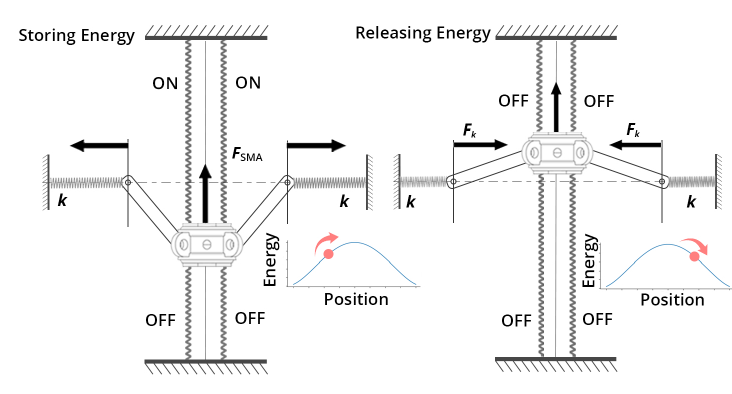}}
    \\
    \caption{Mechanical overview of the DilBot with (a) main body components and (b) concept model of the Bistable SMA-driven engine during actuation of the top pair of springs. Once one pair of SpIRAl actuators reaches the activation temperature, their state switches from OFF to ON, pulling the central hub from an equilibrium position through the central rod with force {\boldmath $F_{SMA}$} while linkages push the $k$ springs, slowly storing energy on them. As the central rod reaches the unstable peak energy position, the $k$ springs quickly release the stored energy with force {\boldmath $F_{k}$}, causing the central hub to snap to reach another equilibrium position. The used actuators return to the OFF state, and the other pair can be readily activated to start the next stroke in the opposite direction.}
  \label{fig:design} 
\end{figure*}

The design of the DilBot aims to generate pulse-jet locomotion similar to the individuals in the Cubozoa class, squeezing water out of a bell periodically to produce the required thrust. We achieve that via the two primary robot components: the encasing silicone bell, pictured along with other various components in Fig. \ref{fig:design}(a), and an internal bistable SMA pulse engine, conceptually pictured in Fig. \ref{fig:design}(b). 

The bell, a hollow paraboloid that surrounds the internal SMA pulse engine of the DilBot, mimics the characteristically elongated bell of a box jellyfish. A mould was modelled and 3D-printed to cast the bell out of near-clear silicone (Ecoflex 00-31, Smooth-On Inc., USA). The bell's thickness is tapered from 2 mm to 1 mm to strengthen the cavity walls and allow for high deformation. The bell extends lengthwise, serving as a passive flap to improve thrust, a feature discussed in \cite{Villaneuva2011}. The silicone bell interfaces to the interior via a segmented outer ring, which is mechanically limited to a radially expanding motion through sets of parallel bar linkages. Along the parallel links, ribs are extruded to provide structural support to the bell and limit the inward collapse of the silicone during the engine power stroke. 

Within the bell, the engine structure is supported by a central rod, rigidly fixed to a cap and lid at the top and to a fixed hub at the bottom. A central hub slides along the central rod between the lid and the fixed bottom. The cap, outer ring, and hubs of the DilBot were printed from resin due to its high-temperature resistance, which was necessary for proximity to heated SMA springs. In contrast, the linkages are laser cut from 2 mm acrylic due to ease of manufacturing and lightweight, enabling various link lengths and designs to be rapidly switched during iterative design and testing. Joints between moving components are formed using snap-in nylon rivets to reduce weight and minimize rust. Relevant design parameters of the DilBot are included in Table \ref{tab:parameters}. 

The force required to deform the silicone bell was empirically determined to be approximately 6 N and was used as a reference when determining the actuator parameters. These parameters were calculated using methods and formulas from \cite{Sung-Min2012}. Each specification was chosen to ensure operation within a repeatable range so as not to wear out or permanently deform the coiled SMA wire. 

We made the SMA springs by winding the SMA wire (Flexinol, DYNALLOY Inc., USA) in a threaded rod and fastening it with nuts at the ends. Following the manufacturer's instructions, each contraption was thermally annealed and then cooled, thus obtaining the coiled SMAs. Each coil was insulated using 0.5 mm ID x 1 mm OD silicone tubing to allow underwater operation. Marine grease and pliers were used to facilitate the insertion process. The tubing segments were then sealed with silicone rubber adhesive (Sil-Poxy, Smooth-On Inc., USA) and the ends were crimped and connected to ring terminals via nylon bolts and washers. The bolted connections are electrical and physical links to the hubs where the springs are mounted. Pairs of SpIRAls are attached between the lid and central hub and between the central hub and fixed bottom hub. Relevant design parameters of each SpIRAl actuator are summarized in Table \ref{tab:parameters}.



\begin{table}
\caption{Design Parameters of the DilBolt Prototype}
\label{tab:parameters}
\centering
\begin{tblr}{
  cell{1}{1} = {c=2}{c},
  cell{1}{4} = {c=2}{c},
  hline{3} = {1-2,4-5}{},
}
\textbf{ SpIRAl Actuator}   &                &  & \textbf{ Robot}    &                              \\
\textbf{Parameter} & \textbf{Value} &  & \textbf{Parameter} & \textbf{Value}               \\
Wire Diameter      & 0.381 mm       &  & Total Mass         & 186.0 g                      \\
Coil Diameter      & 3 mm           &  & Bell Mass          & 64.6 g                       \\
Coil Length        & 40 mm          &  & Body Length        & 228.6 mm                     \\
Maximum Stroke~    & 2.1 mm         &  & Body Diameter      & 110.0 mm                     \\
Pitch              & 0.5 mm         &  & Bell Volume        & 1113.9 cm\textsuperscript{3} \\
Active Turns       & 18             &  &                    &                              
\end{tblr}
\end{table}

\subsection{Bistable SMA-driven engine}
\label{subsec:engine}
What we refer to as an SMA pulse engine, or bistable SMA-driven engine, provides the necessary mechanical power for each swimming stroke of the DilBot. An upstroke is the resulting motion when the top pair of SMA springs is actuated, like in Fig. \ref{fig:design}(b), sliding the central hub through the rod closer to the lid. Consequently, a downstroke results from actuating the bottom pair of SMA springs, moving the central hub closer to the fixed hub. The engine is a composition of mechanical linkages and grouped antagonistic SMA coil actuators that exhibit a bistable mechanical behaviour through both directions of an oscillating power cycle.

\begin{figure}[tb!]
    \centering
    \includegraphics[width=1\linewidth]{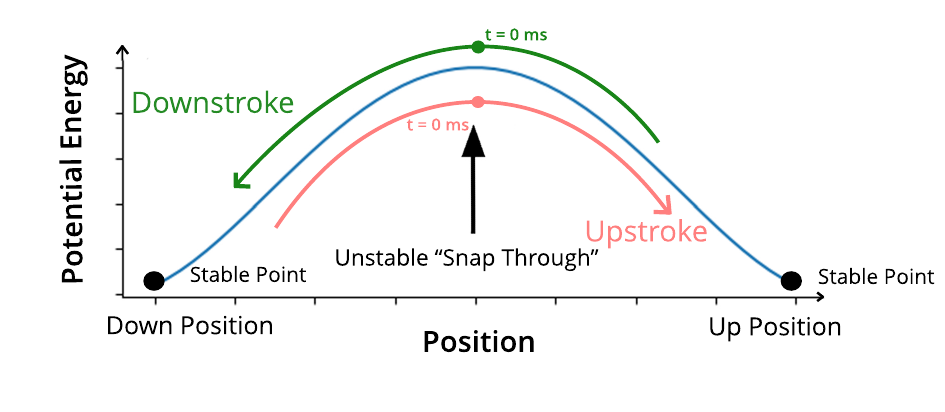}
    \caption{Theoretical energy plot of a bistable system, which presents two stable equilibrium states. The upper and lower curved arrows showcase what happens in terms of energy in our proposed system for both actuation directions, where an unstable central position causes the central hub to ``snap through.''}
    \label{fig:bistability}
\end{figure}

Bistability refers to a system with two stable equilibrium states, separated by some threshold of potential energy presenting an energy diagram similar to the one shown in Fig. \ref{fig:bistability}. In our case, the stable equilibrium points of the cycle occur when the central hub, depicted in  Fig. \ref{fig:design}, lies at rest in either the up or down position. Physical stops placed on the central rod limit the motion of the central hub past these points. The bistability suits the bio-inspired locomotion of box jellyfish, maintaining a stable glide phase at equilibrium points and transitioning in and out of a high potential energy state during propulsion.

Conceptually, the auxiliary springs with elastic constant $k$, depicted in Fig. \ref{fig:design}(b) and located on either side of the engine, serve as a potential energy barrier that separates the two positions into distinct equilibria. Moving from one equilibrium to another involves storing potential energy in the auxiliary springs and releasing it. 

We hypothesized that adjusting the engine to store the elastic energy slowly and then quickly release it would produce an underwater jet that could propel a robot forward. Hence, DilBot's design translates this concept by replacing the auxiliary springs with the silicone bell. Its deformation serves the same potential energy barrier function, allowing consecutive water suction into the bell (bell expansion) and ejection (bell contraction). 

Pairs of SpIRAl actuators need to move the central hub to allow such motion, which is transferred through linkages perpendicularly into the elastic bell.  Typically, SMA coil actuators must be ``reset'' to a stretched position before they can be powered again. However, the actuators in the proposed system are antagonistically configured such that during cycling, one actuator pair is always charging by storing energy for a power stroke. This symmetric configuration of antagonistic SMA coil actuator pairs and the elastic spring force of the bell allow the central hub to ``snap through'' an unstable equilibrium point at the middle of the central rod into either the top or bottom position in a high-power motion. During each transition, the linkage-coupled expanding elements experience a rapid inward contraction, taking advantage of the high-force generating capacity of the SMA actuators to store energy in the bell. This energy is released passively as a result of the bistable structure. 

\subsection{Principles of Locomotion}
\label{subsec:locomotion}

A mathematical model of thrust produced by jellyfish can be borrowed from existing literature and applied to the locomotion of the DilBot, as both produce thrust from the forceful ejection of water and resultant momentum exchange. As discussed in \cite{doi:10.1139/z83-190} and \cite{RoboScallop}, the magnitude of the thrust produced is given by the rate of momentum efflux, which is a function of the change in volume of the bell cavity. Furthermore, for constant rates of momentum efflux and volume change, the thrust becomes primarily a function of the duration of expansion and contraction,


\begin{equation}\label{thurst_momentum}
   T = (dm/dt)\cdot u\textsubscript{e} = (\rho Q)\cdot(\frac{Q}{A})=(\frac{\rho}{A})\cdot(\frac{\Delta V}{\tau})\textsuperscript{2}, 
\end{equation}
where \textit{T} is thrust, ${m}$ is the instantaneous mass including the mass of the water in the cavity,  ${u\textsubscript{e}}$ is the velocity of the ejected fluid, $\rho$ is density, and ${Q}$ is flow rate. The fluid velocity can be described with flow rate over cross-sectional area ${A}$ \cite{RoboScallop}. This area is representative of the velum (jet orifice) of the jellyfish. Additionally, ${Q}$ can be written as the change in volume $\Delta V$ throughout contraction or expansion $\tau$. 
    
From the analysis of \eqref{thurst_momentum}, thrust is inversely proportional to $\tau\textsuperscript{2}$. This relationship between thrust and time duration complements the slow storage of potential energy in the bell and quick ejection of water, which the DilBot implements to locomote. This asymmetry between time durations allows for the net forward momentum of the robot. Not considered are the effects of instantaneous drag and force to overcome the robot's inertia. These forces occur instantaneously during the thrusting period and continue to slow the robot during the glide period. These effects are estimated and discussed along with the results in section \ref{sec:results}.


\section{Experimental Methods}
\label{sec:experiments}

This paper primarily evaluates the feasibility of the proposed SMA-driven engine designed for pulse-jet locomotion in soft aquatic robots. We conducted three distinct types of experiments with the prototype, which are detailed in subsections \ref{subsec:swim}, \ref{subsec:force}, and \ref{subsec:flow}. Throughout testing, the DilBot was powered via a three-wire tether to an external power supply, which provided 15 V and 8 A DC to the SMA coils. Two wires provide power for each set of actuators, while the third one is the common ground. Approximately every five seconds, the power supply was turned on and off manually to allow the coils to cool before the next actuating cycle. Additionally, the water tanks used for the experiments were filled with tap water and left to rest at room temperature at least six hours prior a trial.

\subsection{Free Swimming}
\label{subsec:swim}

Free swimming experiments were conducted to attest to the prototype's locomotion capability. The DilBot was placed inside a 76.2 cm x 47.4 cm x 31.3 cm glass tank full of water and connected through a tether to the power supply. To enable horizontal swimming, we added closed-cell foam pieces to either end of the rod to trim the buoyancy of the DilBot until it barely sank or floated. To reduce the impact of the weight of the tether, a foam platform was placed on the water's surface for the wires to rest on. 

Before actuation began, the robot was placed and allowed to settle at one end of the tank. The power to the springs was alternately applied to each of the positive tether wires to select the direction of actuation required for generating cyclic pulsing. The motion was recorded with a video camera (Cyber-shot DSC-RX100 IV, Sony Group Corporation, Japan) at 30 frames/s, and distance and speed were analyzed in the video annotation tool Kinovea (https://www.kinovea.org/) using the bottom part of the robot as a visual marker for reference. The software was calibrated using the known grid dimensions placed behind the tank.   

\subsection{Thrust Force Characterization}
\label{subsec:force}

\begin{figure}[t!]
    \centering
    \includegraphics[width=1\linewidth]{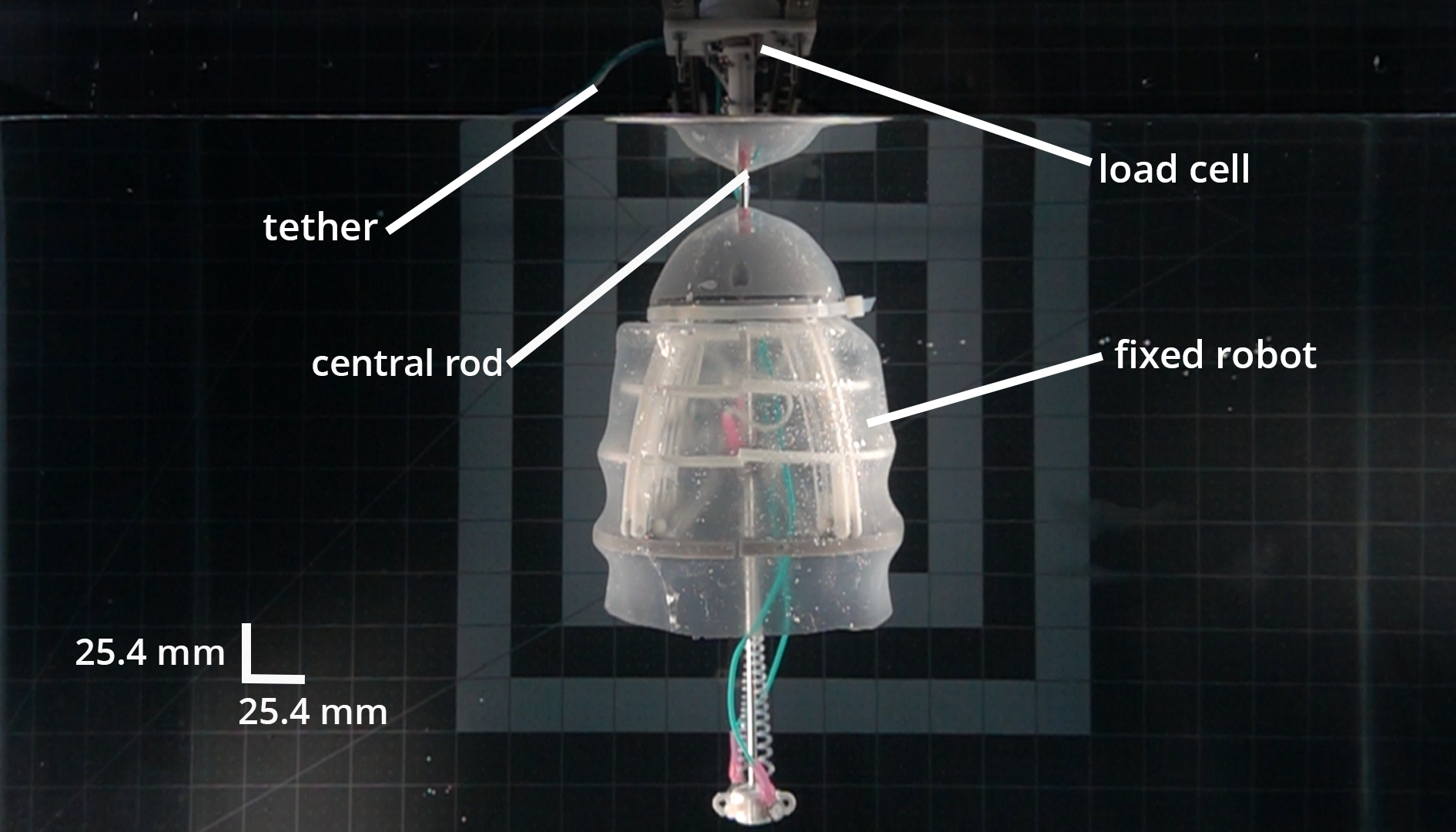}
    \caption{Thrust force characterization experimental setup.}
    \label{fig:force}
\end{figure}

To obtain thrust force measurements from the prototype during upstroke and downstroke, the carbon fibre central rod was switched to a 3 mm x 30 mm steel reinforcing rod mount to provide more rigidity and limit undesired horizontal motion. The new rod was then fixed to a six-axis load cell (Nano 17, ATI Industrial Automation, USA) by a PLA 3D-printed bracket. Another 3D-printed bracket securely attached the load cell to an aluminum extrusion set above the tank to minimize vibrations. This configuration used the same tank and grid as in the free swimming experiment in subsection \ref{subsec:engine}. The setup is shown in Fig. \ref{fig:force}. 

Power was provided cyclically to one pair of actuators at a time, and force measurements were logged at an effective sampling frequency of 500 Hz. Each of the three performed trials was also captured with the same video camera but at 240 frames/s for motion tracking analysis using Kinovea.

\subsection{Flow Visualization}
\label{subsec:flow}
\begin{figure}[t!]
    \centering
    \includegraphics[width=1\linewidth]{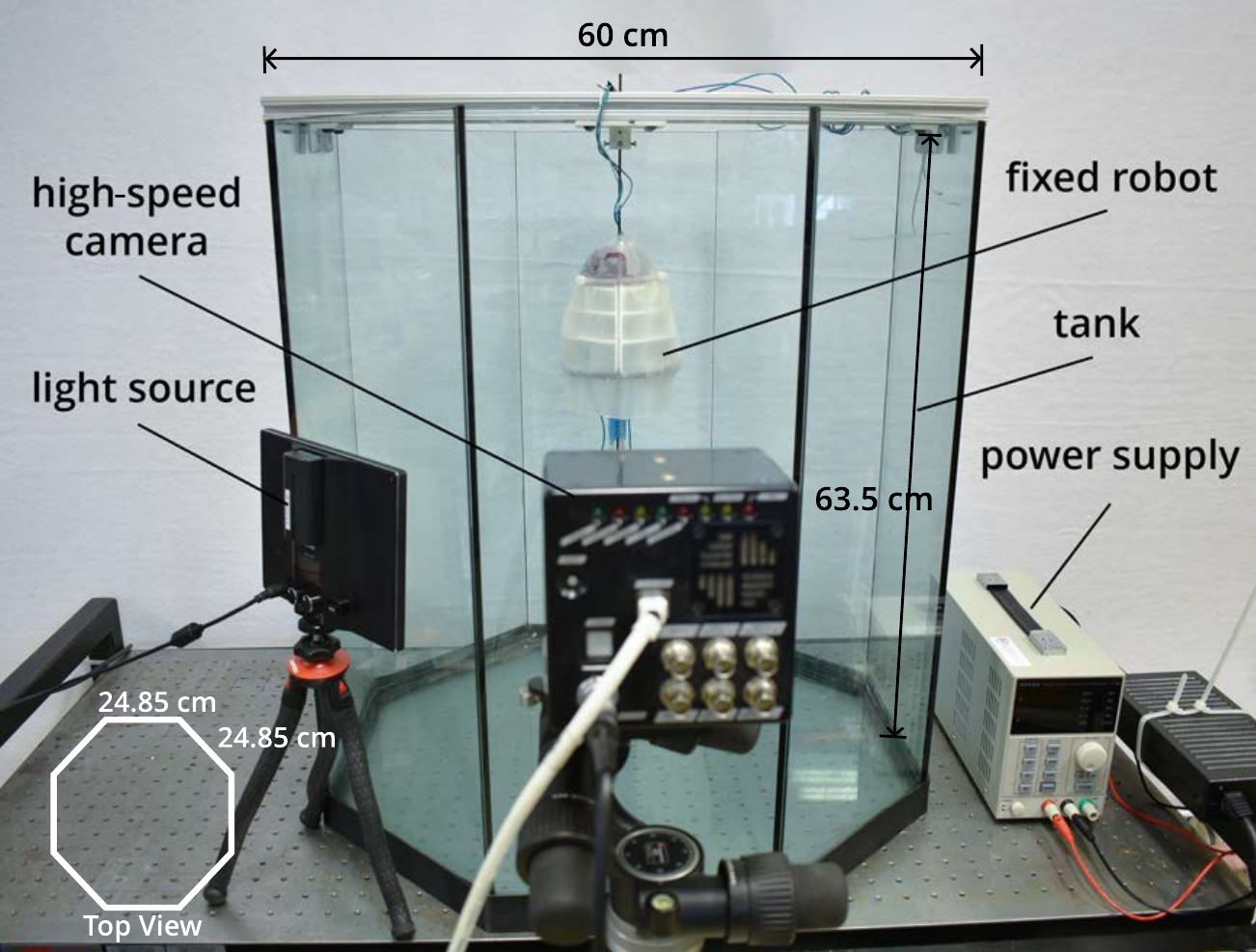}
    \caption{Flow visualization experimental setup.}
    \label{fig:flowsetup}
\end{figure}

The robot was fixed to a 3 mm diameter carbon fibre rod of length  50 mm in a 63.5 cm tall and 60 cm wide symmetric octagonal glass tank full of water to visualize the flow development downstream of the robot during and after spring actuation. Dye was inserted at key positions before actuation, and the resulting flow was recorded on a high-speed camera (FASTCAM Mini WX100, Photron, Japan) at 500 frames/s. LED light sources were placed strategically to clarify and contrast the dye and robot against a white backdrop. The flow visualization setup is shown in Fig. \ref{fig:flowsetup}.

\section{Results and Discussion} 
\label{sec:results}

Following the experimental methods described in section \ref{sec:experiments}, the evaluation of the adopted design approach had the main objective of validating the SMA-driven engine concept. Other influencing aspects, such as water temperature and bell material and shape, as well as their impact on swimming performance, thrust, and rebound dynamics, were outside the scope of this investigation.

\subsection{Free Swimming}

\begin{figure}[b!]
        \centering
    \subfloat[\label{swim_pic}]{%
       \includegraphics[width=1\linewidth]{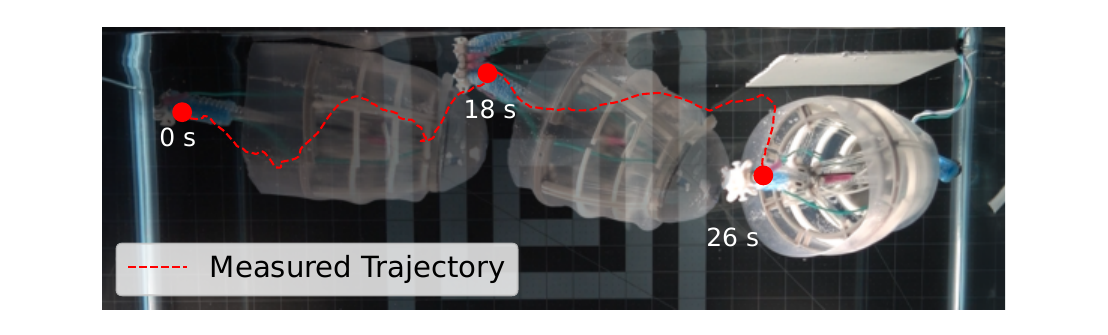}}
        \hfill
    \subfloat[\label{swim_plot}]{%
        \includegraphics[width=1\linewidth]{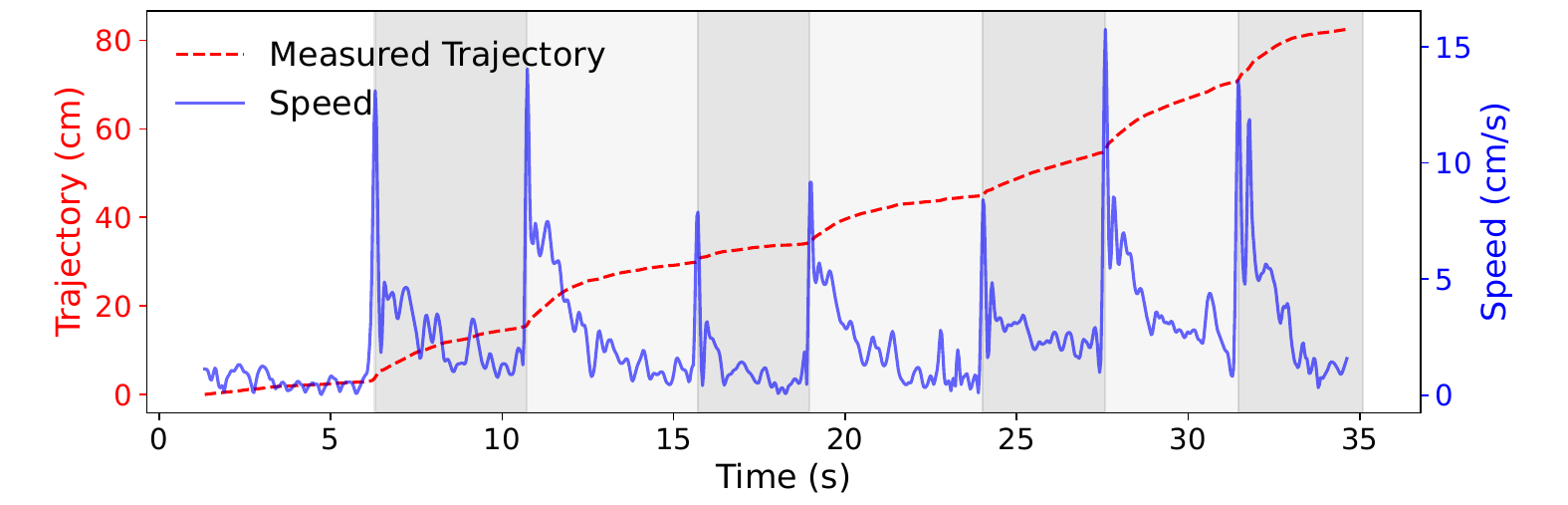}}
    \\
    \caption{Free swimming metrics for the DilBot prototype over seven jet pulses tracked with a visual marker at the bottom of the robot through the video annotation tool Kinovea. a) Measured trajectory, and b) distance and speed in relation to the corresponding swimming stroke. Darker shaded regions correspond to upstrokes, while the brighter ones refer to downstrokes.}
    \label{fig:swim}
\end{figure}

The DilBot's bistable SMA-driven engine successfully generated pulse-jet locomotion underwater. The measured trajectory and speed during the horizontal free swimming experiment are shown in Fig. \ref{fig:swim}. 

We recorded a maximum speed of about 158 mm/s, or 0.7 body length/s (BL/s). However, as expected and indicated by the plot of speed over time in Fig. \ref{fig:swim}(b), this maximum speed is only briefly present due to jellyfish's intrinsic unsteady swimming behaviour. Most of the robot's swim cycle is spent in a slow glide forward, with an average velocity recorded across the free swim test of about 2.74 mm/s or 0.012 BL/s. While most of the time is spent in this slow glide, the POC prototype was not optimized to minimize the drag forces that work against the forward motion. Improvements to the design, including a reduced weight or slimmer profile during contraction, are considerations for future work.

As observed, the swimming trajectory in Fig. \ref{fig:swim}(a) was not linear, resulting in a total distance greater than the width of the tank. This happens partly because of the momentum created by the counterweights and foam pieces added to the structure to make it horizontally buoyant, partly due to the drag inflicted on the robot by the tether and the foam platform. Additionally, since the tracking software only analyzes 2D frames, a component of the robot's motion is not visible when the DilBot swims slightly toward or away from the camera. Consequently, at this stage, the measured trajectory can only be used as an estimation. For the same reason, some measurements for strokes in the same direction appear lower than others in Fig. \ref{fig:swim}(b). It is also noticeable that the downstroke presents slightly greater speed than the upstroke, around 8\%.

\subsection{Thrust Force Characterization}

\begin{figure}[t!]
  \begin{tabular}{cc}
    \includegraphics[scale=0.53,center]{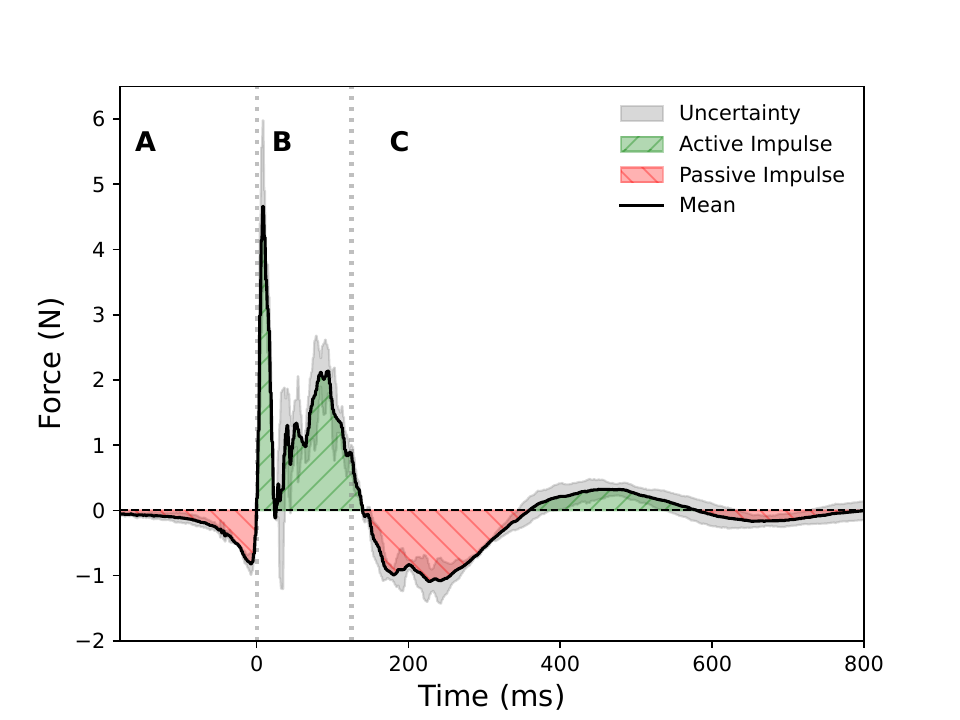} \\(a)\\ \includegraphics[scale=0.53,center]{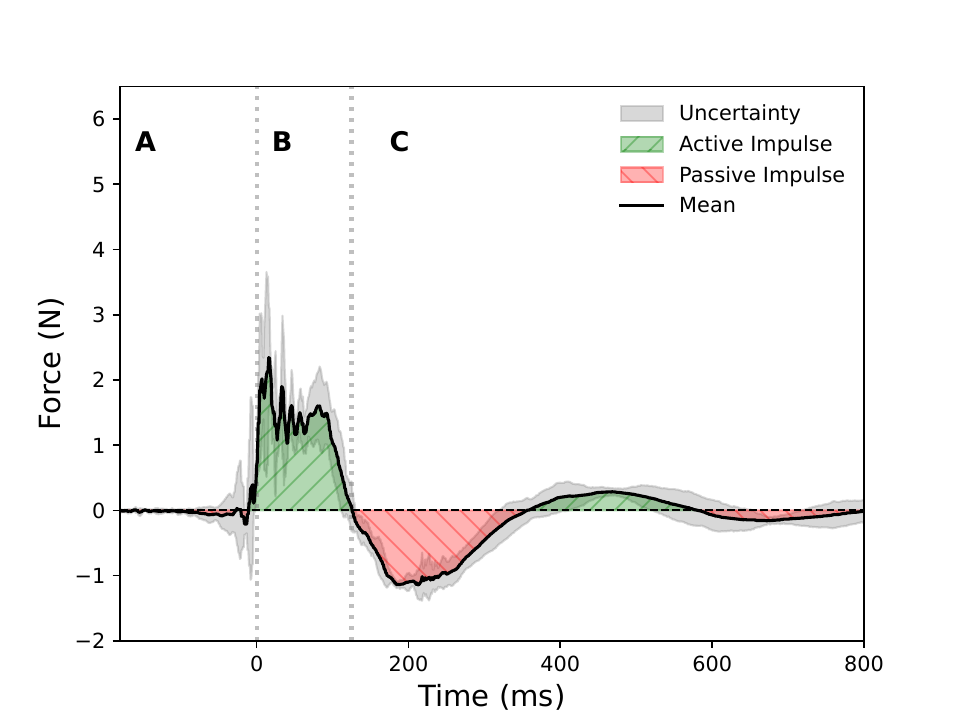}\\ (b)
    \end{tabular}
\caption{Thrust force characterization for the DilBot. A single stroke is divided into A- pre-thrust, B- active thrust and C- rebound. Each graph represents the average of three trials with an uncertainty of two standard deviations for a single (a) downstroke and (b) upstroke. Active impulses designate forces that produce forward motion, while passive impulses ``produce'' drag.}
\label{fig:forces} 
\end{figure}

The same pattern reappears during the force measurements, seen in Fig. \ref{fig:forces}. A maximum peak thrust of 5.59 N was observed during downstrokes, with an average peak thrust of 4.66 N across all trials. A smaller maximum peak of 2.80 N was observed during upstroke trials; the average was 2.60 N. The peaks were used to synchronize the data from each trial, corresponding to the moment of the first contraction so that quantitative analysis could be performed. 

We also divided each stroke into three phases: A- pre-thrust (water being inserted into the bell), B- active thrust (water being ejected from the bell) and C- rebound. We hypothesized the last stage is a repercussion of the combination of the bounciness of the silicone bell, which causes vibration underwater, and the setup for the force characterization experiment, which causes a constraint in the robot. For this reason, the area beneath each force curve in Fig. \ref{fig:forces} is shaded to indicate either an active (green) or passive (red) motion impulse. 

We obtained an average standard deviation across downstroke trials of 0.036 N and 0.38 N for upstroke. The low numbers indicate the robot's high repeatability, especially considering the intrinsic challenges of the underwater environment and the fact that this is simply a POC prototype. The duration and the produced net impulse of each phase are depicted in Table \ref{tab:force}. 

\begin{table}[h!]
\caption{Force Characterization Metrics}
\label{tab:force}
\begin{tabular}{@{}cccc@{}}
\toprule
                                        &                  & \textbf{Downstroke} & \textbf{Upstroke} \\ \midrule
\multirow{2}{*}{\textbf{Pre-Thrust}}    & Duration (ms)    & 200                 & 200               \\
                                        & Net Impulse (Ns) & -0.03               & 0.00                 \\
\multirow{2}{*}{\textbf{Active Thrust}} & Duration (ms)    & 140                 & 135               \\
                                        & Net Impulse (Ns) & 0.18                & 0.15              \\
\multirow{2}{*}{\textbf{Rebound}}       & Duration (ms)    & 660                 & 665               \\
                                        & Net Impulse (Ns) & -0.12               & -0.13             \\ \cmidrule(l){2-4} 
\end{tabular}
\end{table}

Table \ref{tab:force} shows that each stage's alignment of time durations for both strokes is visual and numerical. Considering the uncertainty of two standard deviations for the performed experiments and that since force over time can be related to power, we can also say that the generated energy for each stage is similar by observing the net impulse values. Consequently, both strokes can generate equal positive net impulse, around 0.02 Ns. The positive value indicates that the DilBot generates a similar forward motion for both stroke directions, showing that each stroke imprints the same dynamics on the system and attesting to the symmetry of the proposed mechanism.



\begin{figure*}[t!]
        \centering
    \subfloat[\label{}]{%
       \includegraphics[width=1\linewidth]{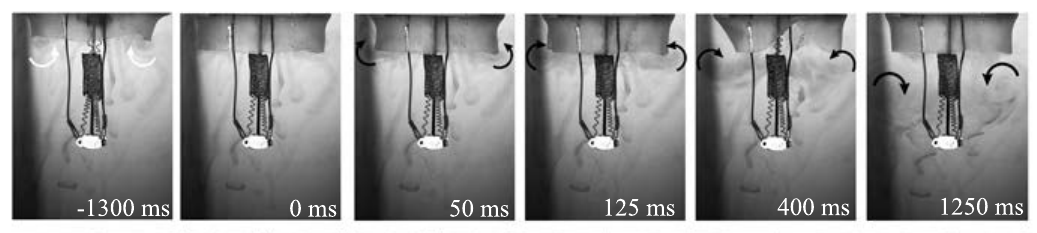}}
        \hfill
    \subfloat[\label{}]{%
        \includegraphics[width=1\linewidth]{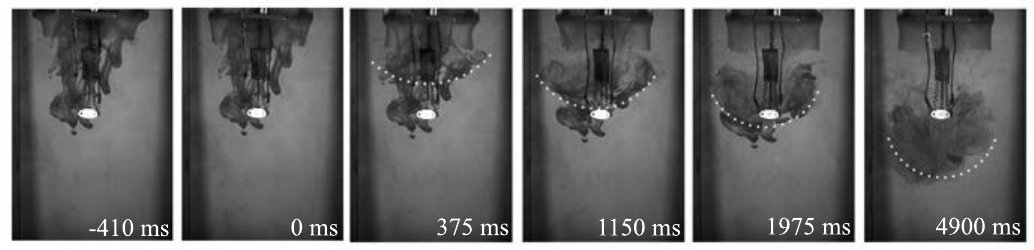}}
    \\
    \caption{Flow visualization results before and after the central hub snaps through the unstable position (0 ms). (a) Zoomed-in view, with the curved arrows showcasing the initial (white, inwards) and propagated (black, outwards) vortices formed; there is no visible vortex in the second frame, and (b) zoomed-out view, with the dotted white line showcasing the jet front and propagated vortices generated by the DilBot; the jet front is not observable yet in the first two frames.}
    \label{fig:flow}
\end{figure*}

\subsection{Flow Visualization}

Flow visualization experiments enabled qualitative assessment of the flow development behind the bio-inspired box jellyfish, providing insight into the flow physics associated with the DilBot's underwater propulsion mechanisms. Fig. \ref{fig:flow} shows frame-by-frame recordings from the high-speed camera. We can see that not only the mechanisms involved in the prototype’s locomotion match the kinematics of a box jellyfish, previously discussed in \cite{shorten2005kinematic}, but also that it can generate flow structures resembling that of natural Cubozoa species \cite{dabiri2006fast,park2015dynamics}. Due to our future end goal of mimicking the energetic locomotion efficiency of jellyfish, our search for generating a similar flow pattern is based on evidence that both initial and propagated vortices serve as an energy-conserving swimming mechanism \cite{han2022dynamic,ruiz2011vortex,kang2023propulsive}. 

In future studies, the qualitative flow visualization results presented herein will be extended with quantitative results from particle image velocimetry (PIV) and laser doppler velocimetry (LDV),  in conjunction with simultaneous force measurements. Due to the complex nature of these experimental techniques, such measurements were beyond the scope of the current work and are planned as future studies. The aforementioned quantitative measurements will facilitate estimates of propulsion strength, vortex strengths, and trajectories, allowing for physical connections to be made between the flow development and jellyfish propulsion.




\section{Conclusion} 
\label{sec:conclusion}

This work introduces a novel approach to underwater locomotion by developing the DilBot, a box jellyfish-inspired POC soft robot. The proposed system utilizes SMA springs arranged in an antagonistic configuration, allowing for efficient energy storage and release. The bistable mechanism is integrated into the silicone bell of the jellyfish robot, which expands slowly and contracts rapidly, providing a powerful propulsion force. The design eliminates the need for a dedicated reset stroke, as the antagonistic springs reset each other naturally. The developed encapsulated SMA springs provided the necessary insulation for underwater operation, enhancing durability and ensuring proper functioning. 

Experimental results demonstrate the feasibility of this design. The robot was able to swim horizontally, reaching a maximum speed of 158 mm/s. It also achieved similar thrust and forward momentum in upstroke and downstroke cycles with low standard deviations across different trials, attesting to the design's repeatability and the engine's actuation symmetry. Flow visualization results enabled qualitative assessment of the flow development behind the bio-inspiration component of the design.

Future work will focus on optimizing the robot’s components and assembly to reduce drag and implementing strategies guided by simultaneous force and quantitative flow measurements to increase control efficiency and minimize energy consumption. Experimentation in environments with different sizes and water temperatures is also necessary to evaluate swimming performance further.

Even with an optimized design for generating high-power thrusts, jellyfish, the organisms we are trying to mimic, are inherently slow swimmers. Integrating the DilBot with bio-inspired behaviours like swarms is desirable for applications where it is necessary to cover large areas quickly, such as environmental monitoring. By approaching this strategy, we aim to build a robot swarm in future work to detect harmful substances and monitor lakes.

\section*{Acknowledgements}

We acknowledge the support of the Natural Sciences and Engineering Research Council of Canada (NSERC), [funding reference number RGPIN-2021-04049] and the Ingenuity Labs Research Institute for partially funding this research.

\bibliographystyle{IEEEtran}
\bibliography{bibliography}

\end{document}